\documentclass{article}

\usepackage[sglblindworkshop, final]{neurips_2026}
\providecommand{\workshoptitle}[1]{}
\makeatletter\providecommand{\@noticestring}{}\makeatother
\workshoptitle{Machine Learning for Systems}
\makeatletter
\renewcommand{\@noticestring}{The 10th Machine Learning for Systems Workshop at NeurIPS 2026, Sydney, Australia.}
\makeatother

\usepackage[utf8]{inputenc}
\usepackage[T1]{fontenc}
\usepackage{url}
\usepackage{booktabs}
\usepackage{multirow}
\usepackage{amsmath}
\usepackage{xcolor}
\usepackage{graphicx}
\usepackage{float}
\usepackage{hyperref}
\usepackage{tikz}
\usetikzlibrary{arrows.meta, positioning, fit, backgrounds}

\makeatletter
\renewcommand{\section}{%
  \@startsection{section}{1}{\z@}%
                {-1.1ex \@plus -0.3ex \@minus -0.2ex}%
                { 0.7ex \@plus  0.2ex \@minus  0.1ex}%
                {\large\bf\raggedright}%
}
\makeatother

\title{Persistent Memory in Multi-Agent LLM Inference:\\
What It Costs, What It Buys, and When You Can Tell}

\makeatletter
\ifdefined\@neuripsyear
\makeatother
\author{%
  Hochan Son$^{1}$\thanks{Lead contributor and corresponding author: \texttt{hochanson@g.ucla.edu}.} \quad Kyungdoe Han$^{2}$ \quad Jaehan Koh$^{3}$ \quad Xiaowu Dai$^{1}$ \\
   Wenlu Xu$^{1}$ \quad Guang Cheng$^{1}$\thanks{Corresponding author: \texttt{guangcheng@stat.ucla.edu}.} \\[6pt]
  \parbox{0.92\textwidth}{\normalfont\small\centering
  $^{1}$University of California, Los Angeles \quad
  $^{2}$University of Wisconsin \quad
  $^{3}$HCLTech America}
}
\else
\makeatother
\author{%
  Hochan Son\thanks{Lead contributor and corresponding author: \texttt{hochanson@g.ucla.edu}.} \\ University of California, Los Angeles
  \and Kyungdoe Han \\ University of Wisconsin
  \and Jaehan Koh \\ HCLTech America
  \and Xiaowu Dai \\ University of California, Los Angeles
  \and Wenlu Xu \\ University of California, Los Angeles
  \and Guang Cheng\thanks{Corresponding author: \texttt{guangcheng@stat.ucla.edu}.} \\ University of California, Los Angeles
}
\fi

\begin{document}

\maketitle

\begin{abstract}
Decomposing long-context inference across cooperating agents bounds the \emph{active} KV
cache per call rather than total evidence, which matters when KV-cache memory binds.
Many such systems add a persistent tier storing and recalling reasoning traces, usually
validated by an ablation reporting an accuracy gain. We measure both on one three-tier agent
architecture. Decomposition delivers: peak KV working set of $14.3$~MiB per query against
$35.5$ and $35.3$~MiB for single-pass and retrieval-augmented baselines. The persistent tier does not: across eight
controlled dataset pairs at $n{=}100$ per arm it costs $+0.368$~MiB~$[+0.167,+0.590]$ of
peak cache and produces no detectable accuracy change ($+0.015$, 95\% CI
$[-0.011,+0.046]$). We argue the null is \emph{structural}: single-question
benchmarks supply each item with its own evidence and score it independently, and
correctness requires resetting stored traces between conditions, so recall has
nothing informative to retrieve. Reaching it took four measurement corrections---three
inflating the apparent benefit, the fourth making an effect that size look resolvable---none
visible in the results table. We give the conditions an agent-memory ablation must satisfy and
detection procedures that need no knowledge of the specific defect.
\end{abstract}

\section{Bounding the working set, and the tier that is hard to test} \label{sec:intro}

The KV cache is the memory bottleneck of long-context inference: it grows with context, is
provisioned against its \emph{peak}, not its average, and routinely exceeds an edge
accelerator's VRAM, which is why cache management is a systems problem in its own right
\citep{kwon2023pagedattention}. One response is architectural: if a task is decomposed
across agents, each handling part of the evidence and passing distilled state onward
\citep{zhang2024chain}, no single call holds the whole context and the \emph{active} working
set stays bounded as total evidence grows.

Systems in this family commonly add a further tier: a persistent store keeping each query's
reasoning trace and retrieving relevant ones later. The argument is intuitive---an agent
that remembers should need less context---and is usually supported by an ablation that
disables the tier and reports an accuracy drop.

We measure both on one such system. The first holds; the second does not---and we argue
the way these systems are evaluated makes it very difficult either for it to hold or for a
reported gain to mean what it appears to.

\begin{figure}[t]
\centering
\includegraphics[width=0.56\textwidth]{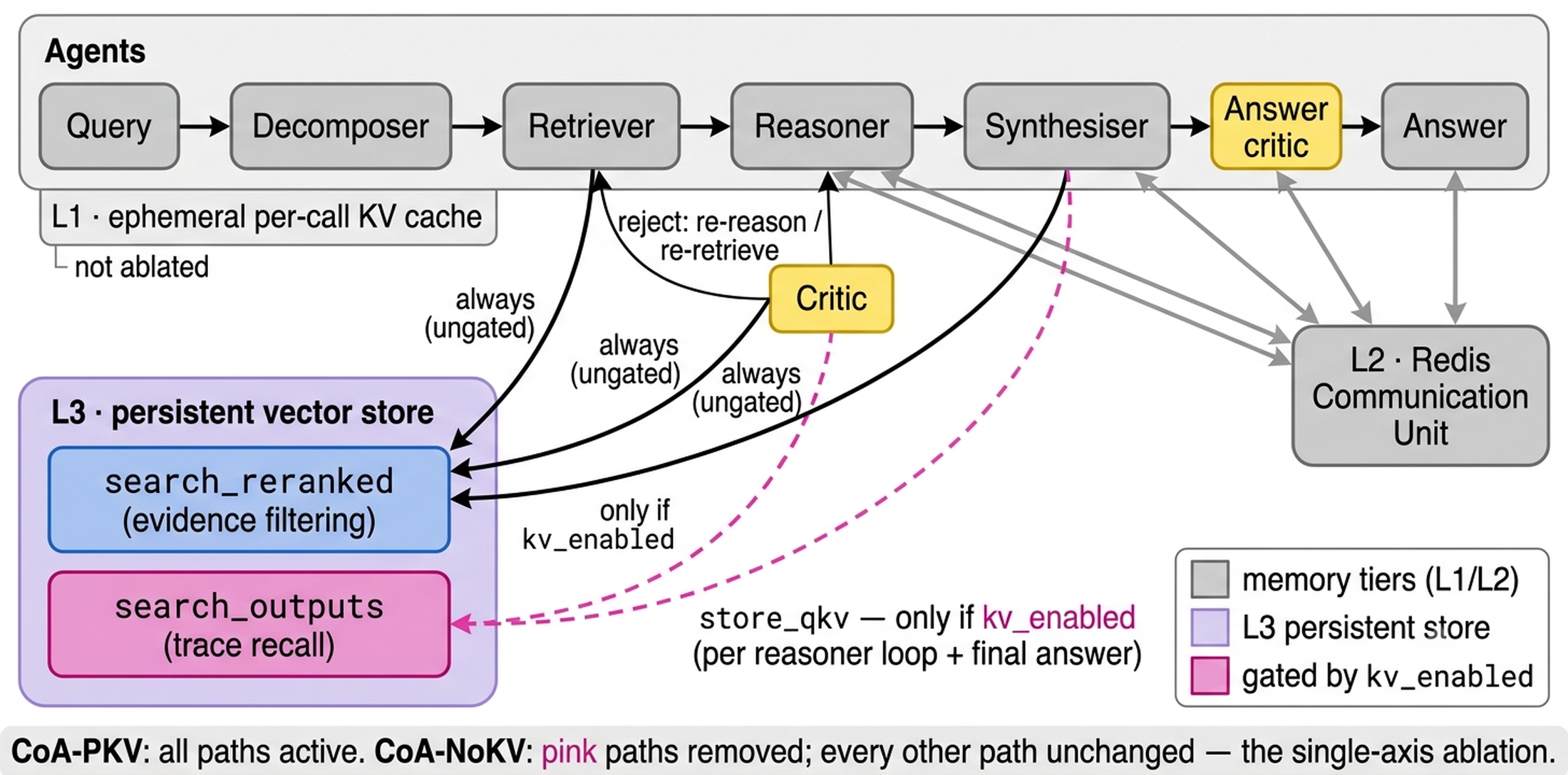}
\caption{The measured system and the single axis the ablation varies; Appendix~\ref{app:repro}
gives the block-level call sequence.}
\label{fig:architecture}
\end{figure}

\paragraph{What is new here.} Retrieval augmentation and agent decomposition are
routinely treated as interchangeable answers to long context. For memory they are
not: RAG changes \emph{which} evidence enters a call; decomposition changes
\emph{how many calls} it is spread across, and only the second bounds the
peak KV. Our retrieval-matched RAG baseline still concatenates its passages into
one generation call and measures a peak indistinguishable from closed-book
single-pass (Table~\ref{tab:working-set}): filtering evidence better does not shrink
the largest call. To our knowledge no prior work reports an agent architecture's peak KV
as a measured, ablated systems quantity (\S\ref{sec:related}); that decomposition alone bounds it,
independent of any persistent-recall tier, is our first contribution and \S\ref{sec:persistent}--\ref{sec:structural}'s null the
second. We do not claim persistent recall improves accuracy: that is untested, not disproven.

\paragraph{The architecture.} Figure~\ref{fig:architecture} shows the system.
\textbf{L1} is the ephemeral per-call KV cache. \textbf{L2} is a key--value store of typed
Communication Units, so an agent receives distilled predecessor output rather than raw
context. \textbf{L3} is a vector store whose two roles---semantic evidence filtering before
anything enters a cache, and $(q,k,v)$ trace persistence for cross-query recall---are
separable in code; only the second is the persistent-memory claim. We call the full system
\textsc{CoA-PKV}---chain of agents over a persistent key--value tier---and the arm with trace
recall and writes disabled \textsc{CoA-NoKV}; they differ in exactly those two operations,
L1, L2 and evidence filtering untouched in both.

\section{The decomposition bounds the working set} \label{sec:decomposition}

We measure peak KV as the largest single call's $(\text{prompt} +
\text{generated})$ footprint, converted at the served model's $10{,}240$\,B/token KV
geometry (not total device memory; Appendix~\ref{app:repro}), against a closed-book single-pass baseline (\textsc{Flat}) and a
retrieval-matched RAG pipeline \citep{lewis2020rag} sharing the same reasoner, critic
and evidence pool. All arms pool the same two cells (HotpotQA, LB-HotpotQA) at
$n{=}100$, averaging each arm's replicate runs (\textsc{RAG} 3 per cell at seeds
42/43/44, \textsc{CoA-PKV} 2, \textsc{Flat} 1); intervals are the Appendix~\ref{app:intervals} cluster
bootstrap.

\begin{table}[H]
\centering
\caption{Peak KV and latency: \textsc{Flat}, \textsc{RAG}, and \textsc{CoA-PKV} (ours). Brackets: 95\% CI.}
\label{tab:working-set}
\small
\setlength{\tabcolsep}{5pt}
\begin{tabular}{lrrrrr}
\toprule
 & \shortstack{\textsc{Flat}$^\ddagger$} & \shortstack{\textsc{RAG}\\\citep{lewis2020rag}} & \shortstack{\textbf{\textsc{CoA-PKV}}\\\textbf{(ours)}} & vs \textsc{Flat} & vs \textsc{RAG} \\
\midrule
peak KV per query (MiB) & 35.5\,{\scriptsize[34.0, 37.0]} & 35.3\,{\scriptsize[30.9, 39.8]} & \textbf{14.3}\,{\scriptsize\textbf{[13.0, 15.4]}} & $\mathbf{-59.9\%}$ & $\mathbf{-59.7\%}$ \\
latency per query (s)$^\dagger$ & 67 & 52 & \textbf{29} & $\mathbf{-56.7\%}$ & $\mathbf{-44.2\%}$ \\
\bottomrule
\end{tabular}
\end{table}

Peak KV is $2.5\times$ lower and latency $1.8$--$2.3\times$ lower: our
\textsc{CoA-PKV} interval clears both baselines with no overlap, while \textsc{Flat} and
\textsc{RAG} overlap each other, as \S\ref{sec:intro} anticipates. \textsc{Flat} has one run per cell, so
its interval covers question sampling only, which by \S\ref{sec:conditions}, C4 makes it
optimistic; $^\ddagger$closed-book single-pass prompting, no retrieval, hence no citation.
$^\dagger$Latency is from one contiguous service window, before the ${\approx}3.5\times$
serving-stack drift of the later campaign (\S\ref{sec:conditions}, C3); peak KV is
token-count-derived and unaffected by serving speed, which is why we lead with it
(per-dataset: Appendix~\ref{app:per-dataset-kv}). Bounded per-call working sets are bought with more
calls---\emph{cumulative} KV traffic is ${\approx}1.6\times$ higher---so the claim is lower
peak at higher total traffic, the trade that matters when KV-cache memory binds. Accuracy against
retrieval-matched RAG is parity, every difference smaller than the reproducibility band;
that the decomposition works is what makes the memory tier worth testing (Appendix~\ref{app:not-reported}).

\section{The persistent tier costs memory and buys nothing measurable} \label{sec:persistent}

We ablate trace recall across eight dataset pairs spanning multi-hop QA
\citep{yang2018hotpotqa,ho2020constructing,trivedi2022musique}, long-document QA
\citep{bai2024longbench} and graduate-level multiple choice
\citep{rein2023gpqa,wang2024mmlupro}. Both arms are matched on
dataset, $n$, seed, evidence scoping and endpoint, traces reset between conditions, and
isolation verified live by a probe asserting the recall path is \emph{present} when enabled
and \emph{absent} when disabled---absence alone is also what an unconfigured store looks
like. Pairing is per question; intervals are percentile bootstrap ($B{=}10^4$), the
across-cell one a cluster bootstrap resampling cells then questions, since questions within
a cell share a difficulty profile.

\begin{table}[H]
\centering
\caption{Cost and accuracy effect of enabling persistent recall.}
\label{tab:recall-effect}
\small
\begin{tabular}{lrr}
\toprule
 & effect & 95\% CI \\
\midrule
peak KV cost of enabling recall & $+0.368$~MiB & $[+0.167, +0.590]$ \\
accuracy change from enabling recall & $+0.015$ & $[-0.011, +0.046]$ \\
\bottomrule
\end{tabular}
\end{table}

The cost interval excludes zero: recall prepends retrieved traces to the reasoner's
prompt, so prompts grow---${\approx}2\%$ against per-query peaks of $13.6$--$21.9$~MiB
(Appendix~\ref{app:per-dataset-kv}), the multiple-choice cells sitting above \S\ref{sec:decomposition}'s $14.3$. The accuracy
interval includes zero. We do not claim the effect is zero; we claim an effect larger than
${\approx}0.046$ is excluded, and that the point estimate is not separable from noise. That
bound is not in tension with the ${\pm}0.09$ replicate band of \S\ref{sec:conditions}, C4:
the band measures run-to-run drift in \emph{absolute} scores, whereas this delta is paired
per question, so drift common to both arms differences out; drift between them, which
C3's order confound leaves open, does not.
It is also convention-dependent: scoring failed questions as $0$ gives $+0.015$,
dropping those either arm failed gives $+0.003$, and the gap is the memory-off arms'
extra failures, which C3's order confound leaves unattributed (\S\ref{sec:conditions}). It is also an intention-to-treat
estimate---recall could fire for only one question in four---which
\S\ref{sec:structural} quantifies.

\section{Why the null is structural} \label{sec:structural}

The result is not that our persistent memory is poor, but that the regime we---and many
systems papers---evaluate in cannot exercise it. That multi-session interaction is required
to test memory is established: LoCoMo \citep{maharana2024locomo} and LongMemEval
\citep{wu2025longmemeval} exist because single-session QA does not probe it. We quantify
that observation's consequence: what a correctly isolated ablation returns when a system is
evaluated on independent-item benchmarks anyway, and what makes it return something else.
Two properties compose to force the null.

\textbf{Items are independent and self-supplied.} Benchmarks of this kind present
each question with its own gold or retrieved evidence and score it in isolation.
Nothing a system learns on question $i$ is intended to be useful for question $j$.

\textbf{Correctness forces the reset.} Recall retrieves by question embedding, so at a
fixed seed a re-run retrieves its own prior conclusions (\S\ref{sec:conditions}, C2).
Preventing that requires clearing stored traces between conditions. But then
within-condition recall can only reach traces from \emph{other, unrelated} items in
the same run.

\paragraph{Measured, not assumed.} Reproducing the retriever's predicate exactly---same
embedding model, same $0.6$ threshold, only traces written earlier in the same run---recall
was reachable for \textbf{24\%} of questions as sent ($20\%$ once each cell's shared
instruction boilerplate is stripped; Appendix~\ref{app:reachability}); for the rest it provably returned nothing,
so \S\ref{sec:persistent}'s figure is intention-to-treat. Restricted to questions where recall could fire the
effect is $-0.012$ $[-0.073,+0.046]$ ($n{=}138$)---a \emph{wider} interval than the
aggregate---while questions where it provably could not act as a built-in control and give
$+0.009$ $[-0.007,+0.026]$, consistent with zero as required. The aggregate is thus correctly measured but
uninformative about efficacy, and the subgroup that tests efficacy is underpowered---a
distinction applying to any such ablation. The transferable point is the contrapositive,
uncomfortable because the practice is common: \textbf{a non-zero memory effect reported on
an independent-item benchmark is more likely an artifact than a finding}. Systems papers
routinely support a memory tier with exactly such an ablation; we did, with four artifacts
(\S\ref{sec:conditions}) present at once---three inflating the effect, the fourth making it
look resolvable. Testing the claim requires a regime where items genuinely share
state---multi-session dialogue, a corpus queried repeatedly, an agent revisiting
entities---and where the reset that prevents contamination does not destroy the signal.

\section{Conditions an agent-memory ablation must satisfy}
\label{sec:conditions}

We reached \S\ref{sec:persistent}'s measurement only after four corrections, stated as conditions rather than
bugs: each was satisfied in appearance and violated in fact. C1--C3 each independently
produced an apparent memory benefit; C4 made effects that size look resolvable.

\textbf{C1 --- The component executed.} Our persistent tier wrote traces nothing ever read:
both arms of every prior memory ablation ran with no memory. Three further components were
similarly inert---a client-side ablation flag for a server-side agent, a reranker whose 404s
became ``keep existing order'', a cost accumulator off the benchmark path that zeroed 38
published cells. Each is \emph{correct} code implementing a degradation that was total,
permanent and indistinguishable from success. \emph{Detection:} emit a counter for every
component whose presence is a claim and assert it non-zero in the run summary---not a log
line, a counter the summary refuses to omit.

\textbf{C2 --- Runs were independent.} With traces keyed by question embedding, a
fixed-seed re-run retrieves its own previous answers; we measured $100/100$ of a run's
questions already carrying a stored trace, from $48{,}915$ traces over $2{,}856$
questions. The signature is detectable retrospectively without instrumentation: within
groups sharing code version, $n$ and seed, score rises \emph{monotonically with run order in
6 of 6 groups} on our most entity-dense dataset ($p\approx0.0006$), with no such pattern
where cross-item entity reuse is lower. Correcting it moved one dataset
$0.2949\rightarrow0.2282$---\emph{downward}, the direction sanity-checking never catches. \emph{Detection:} query the store with the
evaluation questions before running; any hit above threshold is contamination.

\textbf{C3 --- Only the named axis varied.} Our original ablation switch disabled
memory \emph{and} moved execution from a service to in-process \emph{and} changed the
evidence source \emph{and} the dependency set. The cost is concrete: an uncorrected
MMLU-Pro pair read $+0.370$, a $37$-point apparent memory win produced by an option-letter
extraction step fixed in one arm and not the other; with both arms on the same parser the
cell reads $+0.010$. A subtler instance is \textbf{time}: execution was systematically
order-confounded---the memory-enabled arm ran first in all eight pairs---while our serving
stack slowed ${\approx}3.5\times$ across a multi-day campaign. Within a pair the arms ran
at the same speed, but memory-off failed more: 20 orchestrator give-ups to 1, client timeouts
level at 3 (Appendix~\ref{app:per-cell}); with order and arm confounded, the record cannot say why. \emph{Detection:} diff the arms' serialised requests---they must differ in exactly
one field---then counterbalance order, since a diff cannot see time.

\textbf{C4 --- The effect exceeds the measurement floor.} Pooled within-configuration
standard deviation---across replicate groups sharing code version, $n$ and seed---is $0.046$
and $0.052$ on our two multi-hop datasets, giving 95\% bands of $\pm0.09$--$0.10$; every
architecture-vs-baseline delta we had been prepared to quote was smaller. \emph{Detection:}
estimate the band from \emph{replicates}, not a within-run bootstrap, which measures
sampling error only and understated our uncertainty twofold
\citep{bouthillier2021variance}. Under C2's reset protocol the band is far tighter (Appendix~\ref{app:bands}).

\textbf{Why trust a paper reporting four confounds not to have a fifth?} The corrections
moved cells \emph{both} ways---two up, one down, two reverted---the signature of
measurement, not tuning (Appendix~\ref{app:corrections}).

\section{Related work} \label{sec:related}

Agent decomposition \citep{zhang2024chain} and retrieval augmentation
\citep{lewis2020rag} are established responses to context limits; we combine both with a
persistent tier, and both motivate decomposition by accuracy, not memory. KV cache
management \citep{kwon2023pagedattention} optimises the cache for a given workload rather
than asking whether an architectural choice changes its peak demand---complementary to our
question. The agent-memory literature
(\S\ref{sec:structural}) establishes that multi-session regimes are required, not
what happens outside one---our subject. Our bands follow
\citet{bouthillier2021variance}; C2's contamination channel has no counterpart in
single-model benchmarking.

\section{Limitations and Future Work} \label{sec:limitations}

Single system, single team; the conditions are general in aim, not frequency, and
reconstructed rather than derived. Bands rest on 2--4 replicates and \S\ref{sec:decomposition}'s \textsc{Flat}
arm is single-run. \textbf{Future work:} replicate every \S\ref{sec:decomposition} arm to \textsc{RAG}'s three
runs per cell, a context-length scaling study, and \S\ref{sec:structural}'s regime---where agent memory's
accuracy becomes testable, not merely untested.

\newpage
% Scope \small only to the bibliography
{\small
\bibliographystyle{plainnat}
\bibliography{refs}
}

\newpage
\appendix
\normalsize
\raggedbottom  % Prevents LaTeX from stretching vertical gaps when floats move to the next page

% Sections/8-appendix.tex
\section{Appendix}
\label{app:main}
\addcontentsline{toc}{section}{Appendix}

This appendix records what is needed to judge the measurements in the main text. It
is deliberately confined to protocol, statistics, per-cell numbers and the schematic
query flow needed to read them. Agent prompts, the controller implementation and the
typed state schema carried between agents are the subject of a companion paper and
are not reproduced here; Appendix~\ref{app:repro} gives only the block-level structure the
ablation in \S\ref{sec:decomposition}--\ref{sec:structural} already refers to in prose. Per-sample outputs, the isolation
audit, the ablation probe and all interval code will be released with the paper.

\subsection{Reproducibility and implementation}
\label{app:repro}

The controller runs six nodes---Decomposer, Retriever, Reasoner, Critic,
Synthesiser and a separate answer critic---over the three memory tiers named
throughout this paper: \textbf{L1}, a flat, ephemeral per-call context window (not to be
confused with the \textsc{Flat} baseline \emph{arm} in \S\ref{sec:decomposition}, which is an entire
inference pipeline rather than a memory tier); \textbf{L2}, a Redis-backed
key-value store carrying typed Communication Units between agents; and
\textbf{L3}, a Qdrant vector database performing both evidence filtering and,
when enabled, persistent trace recall and write.

Inference is served from two heterogeneous edge accelerators rather than a single
data-center GPU: an NVIDIA Jetson Orin Nano Super (8~GB) running \texttt{llama.cpp}
with a Qwen3.5-2B parameter model, and an NVIDIA DGX Spark running vLLM with
Qwen3.5-35B-A3B-FP8 (35B total parameters, ${\approx}3$B active). Its hybrid
attention keeps a growing KV cache on only 10 of 40 layers, giving the
$2\times10\times2\times256\times1$\,B (fp8) ${=}\,10{,}240$\,B/token used to convert
every KV figure in this paper. Both are on-premises edge devices rather than a cloud
API---the deployment regime the peak-KV argument in \S\ref{sec:decomposition} is meant to address. The Reasoner
and Synthesiser run on the vLLM node; the Decomposer, Retriever, Critic and answer critic
run on the Jetson's smaller model, whose per-token KV footprint is lower. Our accumulator
records the peak call's size but not which agent issued it, so every call is converted at
the vLLM model's geometry. That can only overstate our own peak---a Jetson call scored at
the larger model's geometry is counted heavier than it was---and the \textsc{Flat} and
\textsc{RAG} baselines' single generation call runs on the vLLM node in any case. The
codebase---agents, controller, memory tiers, and the benchmarking and audit tooling
used throughout this paper---will be released as an open-source repository.

\textbf{Peak KV is not device occupancy.} Every peak-KV figure counts the attention cache
of one call, on the 10 full-attention layers above. It leaves out the per-sequence state
the other 30 (linear-attention) layers carry, which the $10{,}240$\,B/token constant does
not include; the rounding of cache allocation to fixed-size pages; the model weights; and
the serving engine's limit on concurrent sequences, which can bind before the KV cache
does. A lower peak KV is therefore headroom in the KV cache, not a proportional gain in
sequences per device.

\begin{center}
\begin{tikzpicture}[
  node distance=4mm, % Compact spacing prevents page overflow
  flownode/.style={draw, rounded corners, minimum width=54mm, minimum height=7mm,
    align=center, font=\small},
  critbox/.style={draw, rounded corners, minimum width=54mm, minimum height=7mm,
    align=center, font=\small, fill=yellow!12},
  lbl/.style={font=\scriptsize\itshape},
  every edge/.style={-{Latex[length=2mm]}, thick}
]
  \node[flownode] (q) {Query};
  \node[flownode, below=of q] (decomp) {Decomposer\\\scriptsize produces sub-questions};
  \node[flownode, below=of decomp] (retr) {Retriever\\\scriptsize \texttt{search\_reranked} (L3)};
  \node[flownode, below=of retr] (rs) {Reasoner\\\scriptsize \texttt{search\_reranked}; \texttt{search\_outputs} recall \emph{if} \texttt{kv\_enabled};\\\scriptsize then \texttt{store\_qkv} \texttt{reasoning\_loop\_}$n$ \emph{if} \texttt{kv\_enabled}};
  \node[critbox, below=of rs] (crit) {Critic\\\scriptsize score vs.\ threshold};
  \node[flownode, below=of crit] (syn) {Synthesiser\\\scriptsize \texttt{search\_reranked}; then \texttt{store\_qkv} \texttt{final\_answer} \emph{if} \texttt{kv\_enabled}};
  \node[critbox, below=of syn] (acrit) {Answer critic};
  \node[flownode, below=of acrit] (ans) {Final answer};

  \draw[-{Latex[length=2mm]}, thick] (q) -- (decomp) (decomp) -- (retr) (retr) -- (rs) (rs) -- (crit);
  \draw[-{Latex[length=2mm]}, thick] (crit) -- node[lbl, right]{accept} (syn);
  \draw[-{Latex[length=2mm]}, thick] (syn) -- (acrit) (acrit) -- (ans);
  \draw[dashed, -{Latex[length=2mm]}, thick] (crit.west) to[bend left=35] node[lbl, left, align=center]{reject:\\re-reason} (rs.west);
  \draw[dashed, -{Latex[length=2mm]}, thick] (crit.east) to[bend right=45] node[lbl, right, align=center]{reject:\\re-retrieve} (retr.east);
\end{tikzpicture}
\end{center}

\textbf{Query path through the controller.} One query's path from arrival to final
answer, drawn from the controller rather than from a narrative summary. Three
details matter for reading the ablation and are easy to get wrong. First,
\texttt{search\_reranked} is not a single step at the Retriever: it runs again
inside the Reasoner and again inside the Synthesiser, and it is \emph{ungated} at
all three. Second, \texttt{store\_qkv} is not a terminal step---it fires once per
Reasoner loop (key \texttt{reasoning\_loop\_}$n$) and once more at synthesis (key
\texttt{final\_answer}), so a query with two Reasoner loops writes three traces, not
one. Third, the Critic is a router, not a gate: on rejection it can return to the
Reasoner \emph{or} back to the Retriever, and a separate answer critic runs after
synthesis. Only the two \texttt{kv\_enabled} operations---\texttt{search\_outputs}
and \texttt{store\_qkv}---differ between \textsc{CoA-PKV} and \textsc{CoA-NoKV};
every other box above executes identically in both arms.

\subsection*{Ablation isolation protocol}
\label{app:isolation}

A cell is admitted only if both arms satisfy all of the following. The checks are
mechanical and are applied by a script over the stored run records, not by
inspection.

\begin{enumerate}\itemsep1pt \parskip0pt
\item The memory-disabled arm was produced by the server-side request field, not by
      any client-side switch. Provenance is recorded per run; an unreadable record
      counts as failing, since guessing ``clean'' is how a confounded run re-enters.
\item Both arms record the same dataset, subset, $n$, seed, evidence-scoping flag and
      service endpoint.
\item The reference arm has memory enabled, so the pair has contrast.
\item Stored traces were cleared before each arm, and the evidence pool was not.
      Clearing evidence would drive retrieval precision toward $1.0$ and collapse the
      axis under study; clearing traces is what prevents the contamination in
      condition~C2.
\item Isolation was verified live in the same window: a probe issues one query with
      memory enabled and one disabled, and requires the recall path to appear in the
      first and be absent from the second. A one-directional check is insufficient,
      because an unconfigured store also produces no recall.
\end{enumerate}

Runs failing any check are reported as gaps rather than numbers. Eight cells pass
and are pooled in Table~\ref{tab:per-cell}; LB-All passes but is reported separately
because it overlaps three of them. The remainder are excluded rather than mixed in.

\subsection{Per-cell results}
\label{app:per-cell}

Paired per question, memory-enabled minus memory-disabled. Each delta carries its own
$n$: accuracy pairs every question both arms scored (failed questions score $0$, so they
are included), while peak KV pairs only questions for which both arms recorded a cache
figure---a failed question has no peak to compare, so the KV $n$ is lower.

\begin{table}[H]
\centering
\caption{Per-cell ablation results.}
\label{tab:per-cell}
\small
\setlength{\tabcolsep}{3pt}
\begin{tabular}{lrrrrrr}
\toprule
cell & $n$ & $\Delta$ peak KV (MiB) & 95\% CI & $n$ & $\Delta$ accuracy & 95\% CI \\
\midrule
HotpotQA            & 96 & $+0.321$ & $[+0.145, +0.504]$ & 100 & $+0.002$ & $[-0.026, +0.029]$ \\
GPQA                & 89 & $+0.640$ & $[+0.083, +1.217]$ & 100 & $+0.090$ & $[+0.010, +0.170]$ \\
MMLU-Pro            & 98 & $+0.748$ & $[+0.314, +1.189]$ & 100 & $+0.010$ & $[-0.060, +0.080]$ \\
LB-HotpotQA         & 95 & $+0.028$ & $[-0.161, +0.226]$ & 100 & $+0.029$ & $[-0.015, +0.077]$ \\
LB-2WikiMultihopQA  & 98 & $+0.221$ & $[-0.018, +0.492]$ & 100 & $+0.007$ & $[-0.031, +0.047]$ \\
LB-MuSiQue          & 99 & $+0.283$ & $[-0.049, +0.635]$ & 100 & $+0.004$ & $[-0.037, +0.045]$ \\
NarrativeQA         & 95 & $+0.071$ & $[-0.198, +0.344]$ & 96 & $+0.004$ & $[-0.027, +0.033]$ \\
Qasper              & 95 & $+0.628$ & $[+0.408, +0.859]$ & 98 & $-0.029$ & $[-0.066, +0.002]$ \\
\midrule
across cells        &    & $+0.368$ & $[+0.167, +0.590]$ &     & $+0.015$ & $[-0.011, +0.046]$ \\
\midrule
\emph{LB-All} (not pooled) & 82 & $-0.007$ & $[-0.236, +0.230]$ & 100 & $+0.071$ & $[+0.017, +0.132]$ \\
\bottomrule
\end{tabular}
\end{table}

\textbf{Why LB-All is reported but not pooled.} It is the harness's combined
LongBench setting, drawing items from the same three subsets that already appear as
LB-HotpotQA, LB-2WikiMultihopQA and LB-MuSiQue. The cluster bootstrap treats cells as
independent units, so pooling it would resample the same questions under two labels
and understate the interval. The exclusion follows from that overlap alone.

It is nonetheless the sharpest illustration of condition~C3 in our data. Its arms
differ by 1 (memory-on) versus 17 (memory-off) failed questions---the largest asymmetry
we recorded. The 17 are connections refused on the run's last questions, during an
orchestrator outage in the second arm; they never reached the pipeline. Since failed
questions score zero, its accuracy delta reads $+0.071$ under our primary convention
and $-0.006$ with every question either arm failed dropped. Had we pooled it, the
across-cell accuracy interval would have shifted to $[+0.001, +0.065]$ and excluded zero:
a marginally significant memory benefit produced entirely by a serving outage during one
arm. We
record this because it is precisely the failure the conditions are meant to catch,
and it would have been easy to accept.

Among pooled cells the GPQA delta is the largest and also carries the most extreme
failure asymmetry (1 memory-on versus 11 memory-off, all orchestrator give-ups); dropping
every question either arm failed reduces it to
$+0.023$, and the across-cell figure from $+0.015$ to $+0.003$. This is the
convention dependence noted in the main text; we report the failures-score-zero
figure as primary because it is the convention under which the arms are scored
identically, and disclose the other because the difference is not negligible.

\textbf{What the record calls a timeout.} The per-question flag both conventions read
marks every failed question, whatever the cause. Over the eight pooled pairs' $794$ paired
questions, memory-on and memory-off fail $7$ and $23$ times: $3$ and $3$ client timeouts
($900$\,s), $1$ and $20$ orchestrator give-ups (HTTP 500, ``Pipeline returned no answer
after 3 attempts'', with no pipeline step or cause recorded), and $3$ and $0$ server disconnects.
The memory-off give-ups all sit in HotpotQA, GPQA and LB-HotpotQA, run on 20~August:
memory-on runs finished 09:08--12:42~UTC, memory-off runs 17:36--21:24. Within every pair the arms ran at
the same speed (median per-question latency ratio, memory-off over memory-on, on
questions both answered: $1.00$--$1.03$), so the campaign's ${\approx}3.5\times$ slowdown
does not explain the gap. With memory-on always first, and on that day also hours
earlier, the record cannot say whether the arm, the order or the time of day caused it;
only a counterbalanced repeat can.

\subsection{Intervals}
\label{app:intervals}

Deltas are paired per question rather than computed from run-level means. Question
difficulty varies far more than the ablated component does, and pairing removes that
variance; matching is on question text rather than row position, since a dropped
sample would otherwise silently pair question $i$ against question $i{+}1$.

Per-cell intervals are percentile bootstrap over the paired deltas, $B=10^4$. The
across-cell interval is a two-stage cluster bootstrap: resample cells with
replacement, then questions with replacement within each drawn cell. Pooling all
questions into one sample would treat several hundred correlated observations as
independent and understate the interval, because questions within a cell share a
dataset, a prompt format and a difficulty profile. The cell is the independent unit.

\textbf{What the paired columns cover.} The two columns of Table~\ref{tab:per-cell} are not
computed over the same set, so each carries its own $n$. Accuracy pairs every question both
arms scored---a failed question scores $0$ and is included, which is the
failures-score-$0$ convention \S\ref{sec:persistent} quotes---so $n$ is $100$ except where a question is absent
from a run entirely. Peak KV pairs only questions for which both arms recorded a cache
figure; a failed question produces no peak, so its KV $n$ is $100$ minus roughly each
cell's failure count. The released \texttt{kv\_layer\_cost\_ci.py} prints the resolved run
ids and both per-cell $n$ values, so the set behind every column is re-derivable rather than
taken on trust.

Peak KV is the largest single call's $(\text{prompt}+\text{generated})$ footprint
within a query, not the sum over calls. The sum answers a different question---total
tokens processed---and is roughly four times larger; using it would overstate the
resource being claimed. Both are recorded per query.

\subsection{Reproducibility bands}
\label{app:bands}

Bands are estimated from replicate groups: runs sharing code version, $n$ and seed,
so that neither code nor sample composition can contribute to the spread. Pooled
within-group standard deviation is $0.046$ across six groups on MuSiQue and $0.052$
across four groups on 2WikiMultihopQA, giving 95\% bands of $\pm0.09$ and $\pm0.10$.

Group sizes are two to four runs, so the bands are themselves imprecise; they are
sufficient to establish that deltas below $0.06$ are not resolvable, which is the use
we make of them, and not to quote to three digits.

These MuSiQue groups serve twice: they estimate the band above and supply
condition~C2's run-order test below. If run order moves scores within a group, the
within-group spread absorbs that movement, so the band is inflated rather than tight---the
conservative direction, given that we use it only to rule deltas unresolvable.

The run-order test in condition~C2 compares each group's scores in chronological order. On MuSiQue all six groups are strictly increasing. With group sizes
$(2,3,3,2,3,2)$ and orderings assumed uniform, the probability of that pattern
arising by chance is $(1/2!)^3(1/3!)^3 \approx 5.8\times10^{-4}$. The comparison
dataset shows one of four, consistent with chance.

\emph{Post-submission measurement.} The bands above are the \emph{historical} spread, from
replicate groups that predate the trace-reset protocol of condition~C2. Re-measuring the same
two datasets under the protocol---four replicates each, fixed seed, traces reset between
runs---gives a within-configuration standard deviation of $0.003$ (MuSiQue) and $0.007$
(2WikiMQA), an order of magnitude tighter, and on a subsequent repair of the serving stack
the four replicates of each were \emph{identical} ($\sigma = 0$). As 95\% bands that is
${\pm}0.006$--$0.014$ rather than ${\pm}0.09$, comfortably inside the ${\approx}0.046$ bound
of \S\ref{sec:persistent}. The floor quoted in C4 is
therefore an upper bound inherited from the contaminated regime, not a property of the
measurement: under the protocol, run-to-run drift on these datasets is not the binding
uncertainty. We report the historical figure because it is what the deltas in this paper
were judged against, and the tighter one because it is what a reader reproducing the work
should expect.

\subsection*{What the corrections cost}\label{app:corrections}

The obvious objection to a paper reporting four confounds is that a fifth may remain.
Our evidence against a systematic residue is that the corrections moved cells in
\emph{both} directions---the signature of measurement rather than tuning. Two
corrections raised scores, one lowered them, and two intended improvements were
regressions we reverted.

\begin{table}[H]
\centering
\caption{What the corrections cost.}
\label{tab:corrections}
\small
\begin{tabular}{lr}
\toprule
correction & effect \\
\midrule
output schema + aligned prompt & $+0.145$ \\
scoped retrieval & $+0.082 / +0.032 / +0.045$ \\
trace-reset protocol & $-0.067$ \\
disabling reasoner CoT (reverted) & $-0.209$ \\
lexical reranker (reverted) & $-4$ of 40 gold-in-top-5 \\
\bottomrule
\end{tabular}
\end{table}

A separate reversal came from sample size alone, with no defect involved: a
comparison reading $0.704 > 0.614$ at $n{=}50$ read $0.622 < 0.646$ at $n{=}100$. We
include it because it is the same category of error---evidence that looked settled
and was not.

\subsection{Recall reachability}
\label{app:reachability}

Reproduces the retriever's own predicate rather than proxying it:
\texttt{search\_outputs} returns stored traces within cosine $0.6$ of the current
question under sentence-transformers/all-MiniLM-L6-v2. Because resets clear the store
between conditions and items are processed in a fixed order, a question can only reach
traces written \emph{earlier in its own run}; we count those. A count of zero means
recall provably returned nothing, which is what makes the unreachable group a control
rather than a comparison. This analysis covers the six multi-hop and multiple-choice
cells it was originally computed on, plus LB-All; NarrativeQA and Qasper, added to the
ablation later, are not re-embedded here. The reachability rate in \S\ref{sec:structural} is over
the six cells ($136$ of $575$ answered questions reachable as-sent); its effect split also
counts LB-All's two reachable questions, giving $n{=}138$. Neither covers all eight cells
in Table~\ref{tab:per-cell}.

\begin{table}[H]
\centering
\caption{Recall reachability by cell.}
\label{tab:reachability}
\small
\begin{tabular}{lrrrr}
\toprule
cell & as-sent & content-only & $\Delta$ reachable & $\Delta$ unreachable \\
\midrule
HotpotQA           & 0\%  & 0\%  & ---      & $-0.006$ \\
GPQA               & 27\% & 24\% & $+0.087$ & $+0.000$ \\
MMLU-Pro           & 86\% & 86\% & $-0.012$ & $+0.000$ \\
LB-HotpotQA        & 2\%  & 2\%  & $+0.000$ & $+0.017$ \\
LB-2WikiMultihopQA & 10\% & 1\%  & $-0.045$ & $+0.019$ \\
LB-MuSiQue         & 16\% & 7\%  & $-0.088$ & $+0.022$ \\
LB-All             & 2\%  & 2\%  & $-0.400$ & $+0.004$ \\
\bottomrule
\end{tabular}
\end{table}

\textbf{As-sent versus content-only.} Multiple-choice datasets wrap every item in
identical instruction text---the stored query for MMLU-Pro is 1{,}499 characters
opening with the same preamble for all 100 items---and that boilerplate carries enough
embedding mass to clear the threshold on its own. Stripping the prefix and suffix
shared by every question in a cell halves the apparent reachability of
LB-2WikiMultihopQA ($10\% \to 1\%$) and LB-MuSiQue ($16\% \to 7\%$): roughly half their
recall was firing on template text. ``Reachable'' should therefore be read as \emph{the
retriever would have returned something}, not as \emph{something useful was available}.
MMLU-Pro is unchanged at $86\%$ because its items genuinely share subject vocabulary,
and it is also the cell where recall fires most and still does not help ($-0.012$).

Two caveats. The reachability figure is threshold- and embedding-dependent; at a looser
threshold more questions become reachable and the subgroup grows. And LB-All's
$-0.400$ rests on two questions---noise, reported only for completeness. Failed
questions are excluded throughout, since they score zero regardless of memory and their
cross-arm asymmetry is confounded with order (Appendix~\ref{app:per-cell}).

\subsection{What is not reported here}
\label{app:not-reported}

Three things are measured but deliberately outside this paper's scope: the full
accuracy comparison against the retrieval-matched baseline, the cumulative-traffic
and throughput characterisation, and the architecture's design rationale. The first
two are summarised in \S\ref{sec:decomposition} only to the extent needed to establish that the
decomposition delivers a bounded working set, which is the premise of the memory
question. A context-length scaling study, which would connect the working-set result
to an asymptotic cost model, has not been run and is stated as a limitation rather
than claimed.

\subsection{Per-dataset KV measurement, apples to apples}
\label{app:per-dataset-kv}

\S\ref{sec:decomposition} quotes one aggregate triple (Flat/RAG/\textsc{CoA-PKV}). This section breaks the
same comparison out per dataset, using the seed-$42$, $n{=}100$ (or $n{=}99$ where one
item failed) slice of the runs behind \S\ref{sec:decomposition}, so every entry below comes from the same
question set run under the same model. These per-dataset figures need not reproduce
\S\ref{sec:decomposition}'s aggregate exactly---\S\ref{sec:decomposition} pools two of these cells and averages each arm's
replicates (\textsc{RAG} over seeds 42/43/44, \textsc{CoA-PKV} over two runs) while
this table is seed-$42$ only---but both are computed identically: peak KV is the
largest single call's $(\text{prompt}+\text{generated})$ footprint, converted at the
served model's $10{,}240$\,B/token KV geometry (Appendix~\ref{app:repro}).

\begin{table}[H]
\centering
\caption{Per-dataset KV measurement, apples to apples.}
\label{tab:per-dataset-kv}
\small
\begin{tabular}{lrrrrrr}
\toprule
dataset & Flat & RAG & \textbf{CoA-PKV} & vs Flat & vs RAG & est/peak$^\ddagger$ \\
 & \multicolumn{3}{c}{peak KV (MiB)} & & & (CoA-PKV) \\
\midrule
HotpotQA           & 35.00 & 30.18 & \textbf{13.59} & $-61.2\%$ & $-55.0\%$ & $12.2\times$ \\
GPQA               & 42.13 & n/a$^\ast$ & \textbf{19.89} & $-52.8\%$ & --- & $8.9\times$ \\
MMLU-Pro           & 36.76 & n/a$^\ast$ & \textbf{21.92} & $-40.4\%$ & --- & $6.3\times$ \\
LB-HotpotQA        & 36.09 & 40.06 & \textbf{15.49} & $-57.1\%$ & $-61.3\%$ & $11.1\times$ \\
LB-2WikiMultihopQA & n/a$^\dagger$ & 38.84 & \textbf{14.34} & --- & $-63.1\%$ & $9.1\times$ \\
LB-MuSiQue         & n/a$^\dagger$ & 44.12 & \textbf{14.55} & --- & $-67.0\%$ & $9.0\times$ \\
\bottomrule
\end{tabular}
\end{table}

$^\ast$RAG does not apply: GPQA and MMLU-Pro are closed-book multiple choice with no
external passage pool to retrieve against \citep{rein2023gpqa,wang2024mmlupro}.
$^\dagger$Not a null result: KV-cost instrumentation was not yet wired into the
Flat backend for these two LongBench subsets at the time these particular runs were
made, and no alternate seed-$42$ Flat run with the instrumentation exists for either.
We report the gap rather than substituting a number from a different seed or a
different dataset. Every reduction above is real regardless: it holds against
whichever baseline (Flat, RAG, or both) is actually instrumented for that dataset.

\textbf{Measured, not estimated.} The $\ddagger$ column is the ratio between each
query's pre-run token estimate---summing $(\text{input}+\text{max-output})$ tokens
declared for every call in the query, before any call executes---and its measured
peak, at the same $10{,}240$\,B/token geometry. For \textsc{CoA-PKV} this ratio is
$6.3$--$12.2\times$: most of the ten-or-so calls a decomposed query makes generate
far below the budget declared for them, and the estimate has no way to know this in
advance, while the peak is whichever single call actually turned out largest. The
same ratio computed identically for Flat and RAG, which make one to two calls per
query, is $1.1$--$1.7\times$ across these same six datasets. Both ratios come from
the same completions that produced the peak and cumulative figures elsewhere in this
paper---not from a separate formula---so the gap between them is direct evidence
that the $2.5\times$ reduction in \S\ref{sec:decomposition} reflects what the served model actually
generated for these questions, not a bound computed before any query ran.

\subsection{Detailed conversation log: good and weak scenarios}
\label{app:logs}

Table~\ref{tab:per-cell} answers whether memory helps on average; it cannot show what
one question's trace looks like, which is where \S\ref{sec:structural} is easiest to misread. This
section works through the only HotpotQA triple sharing seed~$42$ across CoA-PKV,
CoA-NoKV and RAG (r720/r724/r692), so all three answer the identical draw---one
question every arm answers correctly, and one every arm gets wrong.

\paragraph{Good scenario (accept): parity, not a memory effect.} \textbf{Question:}
``My Neighbor Totoro was produced by a Japanese animation film studio founded in
what year?'' \textbf{Gold:} \texttt{1985}. All three arms agree with gold.
CoA-PKV and CoA-NoKV cost nearly identical KV---the recall path found nothing to
recall on this question (a matter of question order and topic, not architecture;
see Appendix~\ref{app:reachability}).

\begin{center}
\begin{tikzpicture}[
  node distance=5mm,
  seqbox/.style={draw, rounded corners, minimum width=115mm, minimum height=8mm,
    align=left, font=\small, inner xsep=4mm},
  every edge/.style={-{Latex[length=2mm]}, thick}
]
  \node[seqbox] (s1) {\textbf{Query} $\to$ \textbf{Decomposer}: ``My Neighbor Totoro was produced by a Japanese animation film\ldots''};
  \node[seqbox, below=of s1] (s2) {\textbf{Decomposer} $\to$ \textbf{L3 (Qdrant)}: retrieve evidence for ``production studio''};
  \node[seqbox, below=of s2] (s3) {\textbf{L3} $\to$ \textbf{Reasoner}: passages (+ recalled trace if \texttt{kv\_enabled})};
  \node[seqbox, below=of s3] (s4) {\textbf{Decomposer} $\to$ \textbf{L3}: retrieve evidence for ``founding year''};
  \node[seqbox, below=of s4] (s5) {\textbf{L3} $\to$ \textbf{Reasoner}: passages (+ recalled trace if \texttt{kv\_enabled})};
  \node[seqbox, below=of s5] (s6) {\textbf{Reasoner}: reason, loop with Critic};
  \node[seqbox, below=of s6] (s7) {\textbf{Reasoner} $\to$ \textbf{Synthesiser}: distilled conclusion};
  \node[seqbox, below=of s7] (s8) {\textbf{Synthesiser} $\to$ \textbf{Query}: \texttt{`1985'}};

  \draw[-{Latex[length=2mm]}, thick] (s1) -- (s2) (s2) -- (s3) (s3) -- (s4) (s4) -- (s5)
        (s5) -- (s6) (s6) -- (s7) (s7) -- (s8);
\end{tikzpicture}
\end{center}

\begin{center}\small
\begin{tabular}{lrrrrrr}
\toprule
arm & answer & $F_1$ & elapsed (s) & calls & peak/cum.\ KV (MiB) & prompt/gen (tok) \\
\midrule
CoA-PKV  & 1985 & 1.00 & 25.8 & 10 & 16.3 / 73.3 & 6{,}624 / 880 \\
CoA-NoKV & 1985 & 1.00 & 25.5 & 10 & 16.8 / 72.7 & 6{,}532 / 908 \\
RAG      & 1985 & 1.00 & 24.5 & 2  & 21.4 / 24.0 & 1{,}154 / 1{,}302 \\
\bottomrule
\end{tabular}
\end{center}

\textbf{Reasoning trace (CoA-PKV and CoA-NoKV are identical on this question).}
\begin{quote}\small
\emph{sub-question: production studio}---``[loop 1] The original question asks for
the founding year of the Japanese animation film studio that produced `My Neighbor
Totoro'. From Evidence [1], I can confirm that `My Neighbor Totoro' was produced by
Studio Ghibli. From Evidence [5], I can find the specific founding details of
Studio Ghibli, which states it was founded on 15 June 1985. Combining these facts,
the studio is Studio Ghibli and its founding year is 1985.''

\emph{sub-question: founding year}---``[answer\_critic loop 1] score=0.00
threshold=0.70 --- needs improvement: Answer is factually incorrect.''
\end{quote}

This second entry is not a fresh answer attempt: it is the answer-critic's verdict
on the first, correct reasoning pass, logged into the same list. The Reasoner
answered both sub-questions correctly in its first pass; the Critic rejected it
once before the pipeline nonetheless emitted the correct final answer
($F_1{=}1$). A critic that scores a correct answer as ``factually incorrect'' and
is overridden downstream is genuine system behaviour, not a rendering artifact, and
is worth its own investigation.

\paragraph{Weak scenario (reject): three different failure mechanisms behind one
metric.} \textbf{Question:} ``Who used a Barrack buster to shoot down a British
Army Lynx helicopter''. \textbf{Gold:} \texttt{IRA}. This is not a memory-ablation
story: all three reasoning traces independently converge on the correct entity,
and each arm then fails for a \emph{different} reason---exactly what a single
$F_1$ number cannot show.

\begin{center}
\begin{tikzpicture}[
  node distance=5mm,
  seqbox/.style={draw, rounded corners, minimum width=115mm, minimum height=8mm,
    align=left, font=\small, inner xsep=4mm},
  every edge/.style={-{Latex[length=2mm]}, thick}
]
  \node[seqbox] (w1) {\textbf{Query} $\to$ \textbf{Decomposer}: ``Who used a Barrack buster to shoot down a British Army Lynx\ldots''};
  \node[seqbox, below=of w1] (w2) {\textbf{Decomposer} $\to$ \textbf{L3 (Qdrant)}: retrieve evidence for ``Who''};
  \node[seqbox, below=of w2] (w3) {\textbf{L3} $\to$ \textbf{Reasoner}: passages (+ recalled trace if \texttt{kv\_enabled})};
  \node[seqbox, below=of w3] (w4) {\textbf{Reasoner}: reason, loop with Critic};
  \node[seqbox, below=of w4] (w5) {\textbf{Reasoner} $\to$ \textbf{Synthesiser}: distilled conclusion};
  \node[seqbox, below=of w5] (w6) {\textbf{Synthesiser} $\to$ \textbf{Query}: \texttt{`Se\'an Kelly'}};

  \draw[-{Latex[length=2mm]}, thick] (w1) -- (w2) (w2) -- (w3) (w3) -- (w4) (w4) -- (w5) (w5) -- (w6);
\end{tikzpicture}
\end{center}

\begin{center}\small
\begin{tabular}{lrrrrrr}
\toprule
arm & answer & $F_1$ & elapsed (s) & calls & peak/cum.\ KV (MiB) & prompt/gen (tok) \\
\midrule
CoA-PKV  & Se\'an Kelly & 0.00 & 31.2 & 10 & 15.6 / 70.5 & 6{,}090 / 1{,}132 \\
CoA-NoKV & Provisional IRA & 0.00 & 33.4 & 10 & 16.8 / 71.2 & 6{,}024 / 1{,}264 \\
RAG      & Provisional IRA & 0.00 & 76.3 & 2  & 46.9 / 49.5 & 931 / 4{,}140 \\
\bottomrule
\end{tabular}
\end{center}

\textbf{Reasoning trace (CoA-PKV and CoA-NoKV are identical on this question).}
\begin{quote}\small
\emph{sub-question: Who}---``[loop 1] The evidence provides specific details about
the incident. Evidence [1] states that on 20 March 1994, a British Army Lynx
helicopter was shot down by the Provisional Irish Republican Army (IRA) in
Northern Ireland. It specifies that a unit of the IRA's South Armagh Brigade fired
an improvised mortar at the British Army base in Crossmaglen. Evidence [5] defines
`Barrack buster' as the colloquial name for several improvised mortars developed in
the 1990s by the engineering group of the Provisional Irish Republican Army (IRA).
Combining these two pieces of evidence, the `improvised mortar' mentioned in the
1994 shootdown incident is the `Barrack buster' described in the definition.
Therefore, the entity that used the Barrack buster to shoot down the helicopter was
the Provisional Irish Republican Army (IRA), specifically the South Armagh Brigade
unit.''
\end{quote}

All three reasoning traces converge on the same correct entity (the IRA) before
answering; the failures happen downstream and are not the same failure.
CoA-PKV's Synthesiser diverges from its own correct reasoning and outputs an
unrelated name (\emph{Se\'an Kelly}). CoA-NoKV and RAG both produce the
semantically right answer, \emph{Provisional Irish Republican Army}, in a surface
form that token-$F_1$ against the gold string \texttt{IRA} scores at zero. One
$F_1{=}0$ collapses three distinct failure modes into the same number, which is
the case for reading per-sample traces before trusting an aggregate.

% Full per-agent reasoning traces, the retrieval evidence, and the script
% (\texttt{scripts/illustrate\_sample\_pipeline.py}) that regenerates this section
% from the underlying run records will be released with the paper.

\end{document}